\documentclass[american, 12pt, a4paper, titlepage]{article}
\usepackage[T1]{fontenc}
\usepackage[latin1]{inputenc}
\usepackage{lmodern}
\usepackage{graphicx}
\usepackage{booktabs}
\usepackage[chatter]{rotating}
\usepackage{fancyvrb}
\usepackage{amssymb,amsmath}
\usepackage{float}
\usepackage{tikz}
\usetikzlibrary{arrows,positioning,shadows,shapes}
\usepackage{subfigure}
\usepackage{stmaryrd}
\usepackage{cite}
\usepackage{xcolor}
\usepackage{amsfonts}
\usepackage{url}
\usepackage{pdfpages}
\usepackage{colortbl}
\usepackage{xcolor}
\usepackage{multirow}
\usepackage{textcomp}
\usepackage{hanging}
\usepackage{ulem}

\begin{document}

\title{Combining {\bf Pr}ediction and {\bf In}terpretation in {\bf D}ecision {\bf T}rees ({\bf PrInDT}) - a Linguistic Example}

\author{
Claus Weihs\\
TU Dortmund University\\
Faculty of Statistics\\
{\normalsize claus.weihs@tu-dortmund.de}
\and
Sarah Buschfeld\\
TU Dortmund University\\
Faculty of Cultural Studies\\
{\normalsize sarah.buschfeld@tu-dortmund.de}
}
\date{\vspace{1cm}\today}

\begin{titlepage}

\maketitle

\end{titlepage}

\noindent{\bf Abstract}\vspace{0.2cm}\\
In this paper, we show that conditional inference trees and ensembles are suitable methods for modeling linguistic variation. As against earlier linguistic applications, however, we claim that their suitability is strongly increased if we combine prediction and interpretation.
To that end, we have developed a statistical method, {\bf PrInDT} ({\bf Pr}ediction and {\bf In}terpretation with {\bf D}ecision {\bf T}rees), which we introduce and discuss in the present paper.

\section{Introduction}

Studies in linguistics are often characterized by small and unbalanced data sets.
This is particularly true for quantitative sociolinguistic studies that investigate the use of non-standard speech forms since such forms often occur at much lower token frequencies than standard speech forms (e.g.\ Buschfeld 2020).
The present paper outlines ways to statistically meet the problems of low token frequencies and unbalanced data sets.
Drawing on child data from Singapore and England, collected by one of the authors, we discuss a new resampling method which builds on the traditional method of repeated random undersampling but enhances it in two important respects.
We evaluate the outcome of undersampling by means of two criteria, i.e.\ interpretability of the results and predictive power.
We therefore assess potential models from both their linguistic and statistical perspectives, which, to our best knowledge, is so far an unprecedented approach.

In this paper, we first introduce the linguistic study, i.e.\ its data and research questions, in Section~\ref{sec:2}. Section~\ref{sec:3} outlines the general statistical approach we take, i.e.\ classification rules and, in particular, conditional inference trees, and discusses their power and value against the background of notions such as prediction and interpretation, unbalanced classes, and re- and undersampling. In Section~\ref{sec:int}, we conflate the statistical and linguistic perspectives. We show how and why interpretability and predictive power are crucial criteria for linguistic studies and how these criteria can be jointly utilized. To this end, we introduce our newly developed statistical approach {\bf PrInDT}. The results of our study are presented and discussed in Section~\ref{sec:5} on the basis of the three best trees generated by {\bf PrInDT}. Last but not least, we look into ensembles of our {\bf PrInDT}-generated trees and show how they can add important information to the findings of individual trees. We finish with some general statements that point the way ahead for both statistical and linguistic research.

\section{Linguistic Hypotheses and Data}\label{sec:2}

The data were collected for a large-scale research project investigating the acquisition of English as a first language (L1) in Singapore (Buschfeld 2020).
Singapore English (SingE) is one of the most extensively researched second language (L2) varieties of English.
However, for an ever-increasing number of children, it is (one of) their home language(s). According to the Straits Times, one of the leading newspapers in Singapore, the number of primary one students who speak mostly English at home has risen to around 70\% in all three major ethnic groups in Singapore, i.e.\ the Chinese, Indian, and Malay (Chan, 2020).
To place L1 SingE on the map of L1 varieties of English and investigate the linguistic characteristics of SingE as well as the acquisitional route Singaporean children take in their linguistic development, we compare the data to data collected from monolingual and bi-/multilingual children in England.
The relevant hypotheses for the study on subject pronoun realization read as follows (Buschfeld 2020, 85):
\begin{itemize}
\itemsep0pt
\item[]{\bf Hypothesis 1}: All children (from Singapore and from England) drop subject pronouns in their early acquisitional stages. The amount of zero subjects is higher in the Singapore group due to the differences in input (SingE behaves similar to null-subject languages, at least at a surface level).
\item[]{\bf Hypothesis 2}: In later acquisitional stages, the Singaporean children drop subject pronouns at a much higher rate than children acquiring English in a traditional native English setting. The children from England ultimately realize subjects as obligatory sentence constituents.
\end{itemize}
To sum up our hypotheses, we expect to find both age-related and ethnicity-related differences for the realization of subject pronouns.

The data were elicited systematically in video-recorded task-directed dialogue between the researcher and the children, consisting of several parts: a grammar elicitation task, a story retelling task, elicited narratives, and free interaction. The recorded material was orthographically transcribed and manually coded for the realization of subject pronouns (realized vs. zero):

The following {\bf examples} illustrate the respective SingE variants (Buschfeld 2020, 144):
\begin{itemize}
\itemsep0pt
\item[1.]	Researcher: [\ldots ] what do you do with your friends? Do you play with them?\\
Child: [ {\bf \o} I] Play with them. Sometimes drawing. [\ldots]\\
Child: Sometimes [ {\bf \o} WE] play some fun things.
\item[2.]	Child: I think in MH370, I think they can find because [ {\bf \o} IT] is easy to go there [\ldots].
\end{itemize}

\noindent The forms in Examples 1 and 2 are variably used by children acquiring English as an L1 in Singapore. They are also typical for adult SingE and thus part of the input the children receive.

The aim of our analysis is to find prediction rules for the use of subject pronouns (realized vs. zero) by means of extra- and intralinguistic variables.
The {\bf extralinguistic variables} considered as independent variables in the statistical analysis are \textsc{ethnicity} (ETH), \textsc{age} (AGE), \textsc{sex} (SEX), \textsc{linguistic background} (LiBa), and \textsc{mean length of utterance} (MLU). \textsc{pronoun} (PRN) is taken into account as the {\bf intralinguistic variable}.
We aim to determine whether any of these variables has a statistically significant influence on the results.
All in all, we extracted 6146 tokens of the values of the subject pronoun variable, each with the full set of extra- and intra-linguistic variables. 528 of these are realized as zero pronouns.

\section{Statistical Modeling}\label{sec:3}

Statistical modeling has long found its way into quantitative linguistics, but still, we would like to first address one of the central questions raised by such endeavors:
What can we expect from statistical modeling?

From a statistical perspective, the answer is quite straightforward. Statistical analyses can yield two kinds of insights. Description elicits information from the data sample. Inference generalizes from a sample to a population.
Both description and inference aim at the {\bf interpretation} of properties of either the observed data (description) or of a more general population (inference).
Inference often comprises the {\bf prediction} of such properties for the more general population.

In order to approximate the relationships between a class variable (e.g.\ pronoun type with possible values `zero' and `realized') and influential variables like extra- and intra-linguistic variables, {\bf classification rules} can be employed. Such rules can be used either for description or inference.
Note that models never depict reality, but are only a hopefully adequate approximation thereof.
To construct such models, we need \, n \, observations (tokens) comprising a value of the class variable and of all influential variables ({\bf learning sample}).

\subsection{Conditional Inference Trees}\label{page:lowfit}

A variety of different methods exists for constructing classification rules.
In the present paper, we concentrate on so-called {\bf decision trees}, which
combine different decisions on individual variables into a set of rules.
The type of tree most often used in linguistics is called {\bf conditional inference tree} ({\bf c-tree}) (cf., e.g., Tagliamonte/Baayen, 2012; Gries, 2020). Such trees include only those decisions that significantly improve the correct prediction about the realization of the dependent variable and employ statistical tests for inference. Therefore, such trees are geared towards generalization / prediction.

\begin{figure}[h]
\centering
\vspace{-0.6cm}
\includegraphics[scale=0.4]{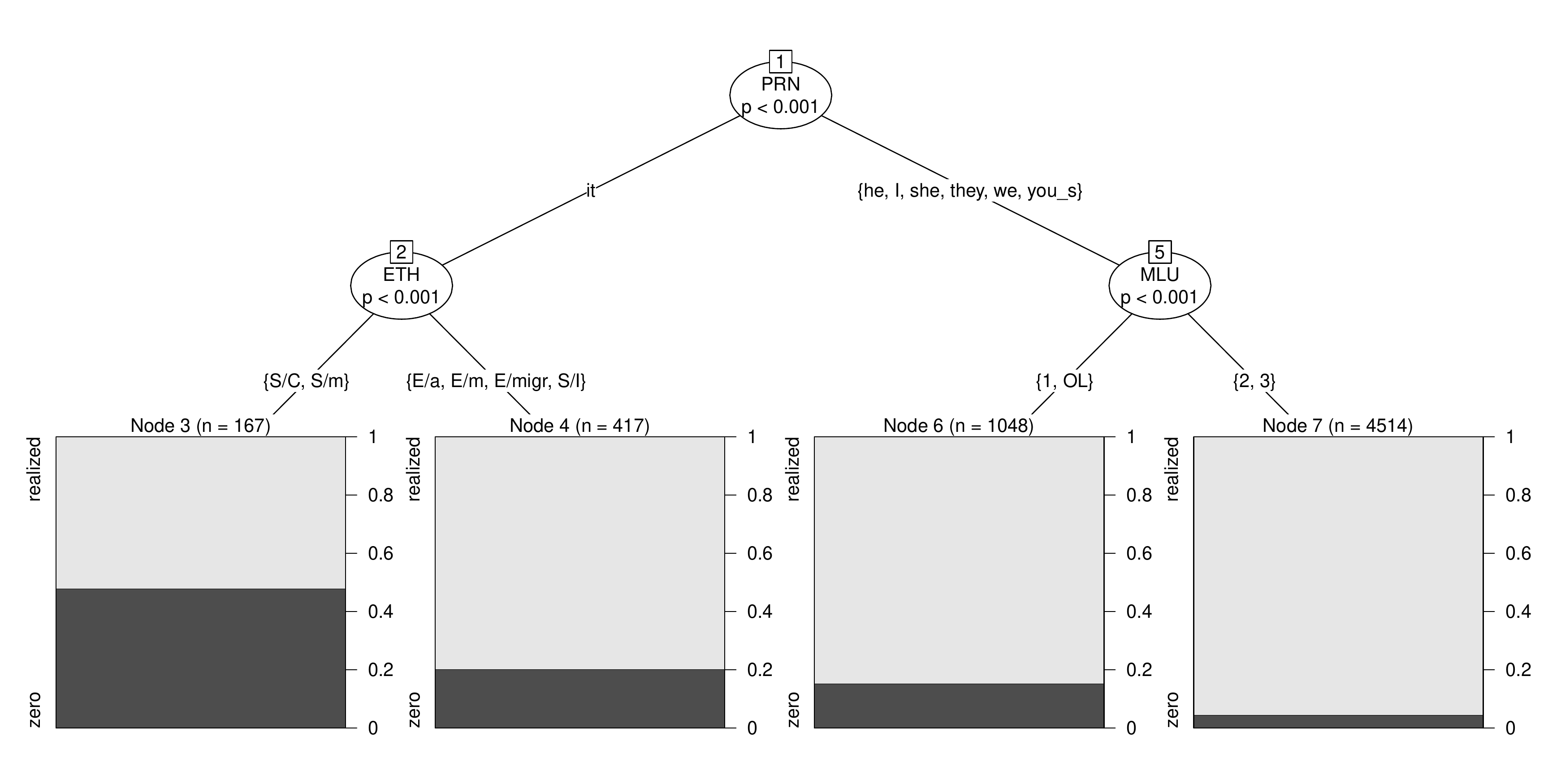}\\
\raggedright prediction: realized \hspace{0.9cm} realized \hspace{1.7cm} realized \hspace{1.65cm} realized\\
\caption{Simple c-tree for pronoun realization.}
\label{fig:simple}
\end{figure}

An example of a c-tree for subject realization can be found in Figure~\ref{fig:simple}.
The tree has to be interpreted as follows: decision variables and p-values of the individual decisions are presented in the `node' corresponding to the decision. e.g.\ PRN and p < 0.001 in node 1. The terminal nodes indicate the relative frequency of each class in the node, e.g.\ 20\% of zeros and 80\% of realized subject pronouns in node 4. The class with the highest frequency is predicted for each token assigned to the node. This is of particular interest for tokens not used for rule construction, i.e.\ {\bf prediction}. Note that the tree in Figure~\ref{fig:simple} predicts only one of the two classes, namely `realized', i.e.\ the zero class is never predicted. The statistical repercussions and approaches to meet this problem are discussed in the following section.

\subsection{Evaluating trees with unbalanced classes}\label{subsec:eval}

Evaluation of classification rules is indispensable to assess their suitability.
The standard evaluation measure for classification rules is\\

\hspace{0.6cm}{\bf overall accuracy} = (no. of correct predictions) / (no. of tokens) .\\

\noindent For the tree in Figure~\ref{fig:simple}, the overall accuracy is 0.914, i.e.\ very high. This is caused by the strong imbalance between the two classes. The zero class is extremely small (528 tokens) whereas the larger class contains more than ten times as many tokens (5618). Although the smaller class is never predicted, the accuracy is very high but clearly misleading. Two things can be done to improve the predictive power of the model: the model quality has to be assessed in a different way and the high imbalance between the classes has to be resolved (cf. Section~\ref{subsec:under}).
For unbalanced classes model quality can be assessed more adequately:
\begin{equation*}
\begin{split}
\text{\bf balanced accuracy} & =  \text{mean of accuracies in the two classes} \\
& =  {(\text{accuracy(zero)} + \text{accuracy(realized)})/2} .
\end{split}
\end{equation*}
\noindent For our example, the value of the {balanced accuracy} = ({\bf 0} + 1.0)/2 = {\bf 0.5} confirms that the high overall accuracy is misleading, since the average of the accuracies of the two classes amounts to only 50\%. The low balanced accuracy and the fact that the smaller class is totally ignored in the predictions render the tree unsuitable for linguistic application.

\subsection{Resampling}

The suitability of a tree-like classification rule strongly depends on whether it is to be used for description or prediction. Do we only want to describe the data set at hand or do we want to predict the behavior of an unknown subject?

If classification rules are used for prediction, their {\bf predictive power} has to be assessed. In order to simulate prediction, we divide our learning sample (i.e.\ the full data set) into training and test sets. This is called {\bf resampling}. Different resampling methods exist. In this paper, we utilize two resampling methods: hold-out and subsampling.

The simplest way of resampling is called {\bf hold-out}, where one part of the observations (say 1/3) is randomly withheld from rule construction (training). Subsequently, a rule is constructed on the training set. This rule is then tested on the hold-out and the test accuracy is used for evaluation. The disadvantage of this method is that the hold-out is generated only once and may not be fully representative of the data set.

{\bf Subsampling} solves this problem by repeating this procedure B times (e.g., B = 200). Then, for each observation (token) the class most often predicted by the B different rules is compared to the observed value of the class variable in the actual data set. The resulting accuracy is used as the evaluation measure for the predictive power of all the B trees generated in the subsampling process.
In the following, we introduce a subsampling method which meets the problem of unbalanced classes.

\subsection{Undersampling}\label{subsec:under}

In order to construct a classification rule with acceptable balanced accuracy, we stochastically reduce the larger class, i.e.\ we apply the method of undersampling.
{\bf Undersampling} takes, in the simplest case, the full sample of the smaller class together with a small sample of the larger class for constructing a rule (training). This procedure is repeated B times (cf. Weiss 2004).

For the {\bf evaluation of undersampling}, we combine goodness of fit and predictive power. The idea is to compute the overall balanced accuracy on all observations of both classes.
For the smaller class, the accuracy is computed on the training sample (fit),
i.e.\ the full data sample of the small class.
For the larger class, it is computed not
only on the training sample (fit), but also on the hold-out from rule construction (prediction).
This way, rules from different training samples can be adequately compared and the best
rule can be identified by looking for the highest balanced accuracy.

In our example, the smaller class (zero) comprises 528 tokens. For undersampling, we decided to subsample the larger class for training. To create a data set with roughly equally-sized classes, we randomly selected 9\% of the larger class, i.e.\ we kept 506 tokens for training. We repeated the process of random subsampling B = 1001 times.

\section{Statistics meets Linguistics}\label{sec:int}

\subsection{Limits on interpretability}\label{page:uninter}

\begin{figure}[thp]
\centering
\hspace*{-0.5cm}\includegraphics[angle=90,scale=0.5]{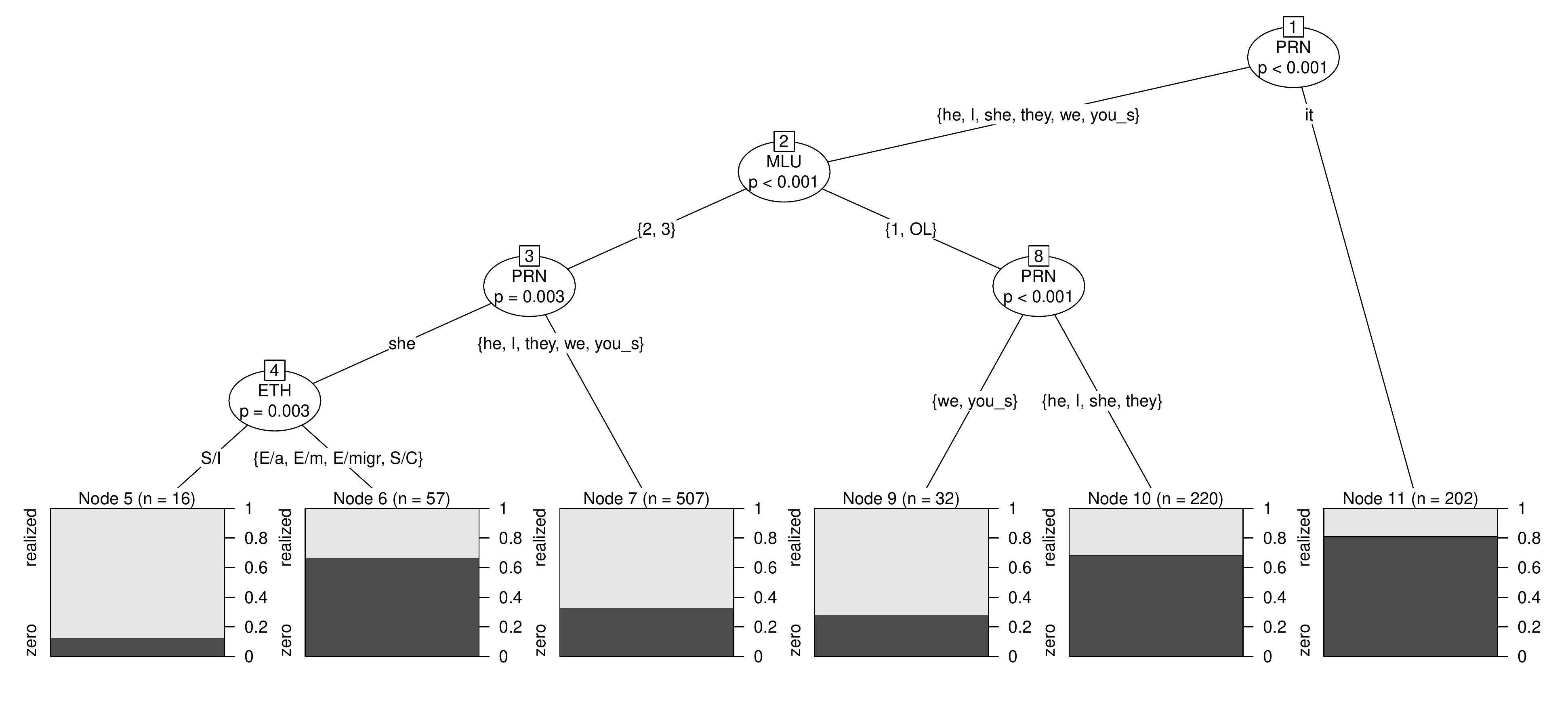}\\
\caption{Uninterpretable tree from undersampling.}
\label{fig:uninter}
\end{figure}

Unfortunately, undersampling might produce trees with high accuracies that are linguistically uninterpretable.
Figure~\ref{fig:uninter} shows such a tree with a balanced accuracy of 0.6898. We later see that this balanced accuracy is only marginally (< 0.02) lower than the best interpretable tree we found (Figure~\ref{fig:best}). The problem in the uninterpretable tree in Figure~\ref{fig:uninter} is the split in node 4. Here, the Singapore Chinese (\textit{S/C}) cluster with the ancestral English (\textit{E/a}) children. This contradicts the linguistic, typologically motivated expectation
to find differences in pronoun realization between these two
groups. First of all, the Chinese languages the Singapore Chinese
children speak as additional languages all allow for zero subjects and it is widely
accepted that languages acquired bilingually influence each other
structurally. Therefore, the Chinese children have a stronger inclination towards zero
subjects than monolingual children growing up in England.
For the Indian children, the linguistic situation is not as clear since the Indian languages spoken by the children do not as unambiguously prefer the zero subject option as the Chinese languages. If any significant differences are to be found between the
groups, these clearly have to be found between the Chinese children and the
monolingual ancestral English ones. Therefore, a combination of the two ethnicities in the same group of values has to be excluded.

To make sure that the resulting tree is linguistically interpretable, we have identified the following combinations of values of our variables as not permissible in splits of interpretable trees:\\

\noindent {\bf Excluded groups of values in our linguistic example}: \\
ETH == \{E/a, S/C\},\\ETH == \{E/a, E/m, S/C\},\\ ETH == \{E/a, E/migr, S/C\},\\
ETH == \{E/a, E/m, E/migr, S/C\},\\
ETH == \{E/a, E/m, E/migr, S/C, S/I\},\\ ETH == \{E/a, E/m, E/migr, S/C, S/m\},\\
ETH == \{E/a, E/m, S/C, S/I\},\\ ETH == \{E/a, E/migr, S/C, S/I\},\\
ETH == \{E/a, S/C, S/I\},\\ MLU == \{1, 3\}\\

\noindent The above restrictions mainly concern \textsc{ethnicity} excluding all combinations entailing \{E/a, S/C\} for the reasons stated above. Moreover, splits with MLU are restricted by excluding the combination of MLU groups 1 and 3. The decision to exclude this combination is again linguistically motivated. Group 1 contains children of 45 months and younger. Language acquisition research has shown that in this age group, children often produce subjectless sentences, not only in Singapore but also in England (e.g. Roeper \& Rohrbacher 2000). The children in group 3 are all 7 years and older. The children from England in this group have thus clearly moved beyond the zero-subject stage. The children from Singapore still use zero subjects, due to the reasons outlined earlier, but to a lesser extent. This is also due to the decreasing acquisition effect but also since they enter formal schooling and thus more standard language exposure from age 7 onwards. Therefore, the children in MLS groups 1 and 3 show completely different inclinations towards zero subjects.

\subsection{Interpretability meets predictive power}

On the basis of the above deliberations, we are able to generate trees with both predictive power and interpretability if the following criteria are met:
\begin{itemize}
\itemsep0pt
\item[] {\bf Interpretability}: no excluded grouping is included in the tree.
\item[] {\bf Predictive power}: the balanced accuracy is `high enough', e.g.\ greater than a threshold $c$.
\end{itemize}

\noindent We propose two kinds of tree-like approaches that meet these criteria. As a first step, we use that tree from undersampling which is interpretable and has the highest balanced accuracy. As a second step, we employ {\bf ensembles of rules} that meet the criteria. Such ensembles are defined in the following way:

The B rules from undersampling are assessed by the two criteria defined above.
To assess predictive power, in our example, the threshold $c$ is set to the value of the median of balanced accuracies in undersampling.
All rules from undersampling meeting the two criteria are combined to a so-called {\bf ensemble} of rules.
For each observation, the ensemble of rules predicts that class which is most often `predicted' by the individual trees in the ensemble. These predictions are the basis for calculating balanced accuracies.

\section{Results}\label{sec:5}

All results were created by means of the software R (R Core Team 2019). The decision trees were generated by the R-package `party'. Note that, since the significance limit of 0.01 is used, the trees are smaller than for the standard limit of 0.05. This facilitates straightforward interpretability in order to illustrate our line of argumentation. The following results are generated by means of the R-function PrInDT, that implements our newly developed procedure.\footnote{The source code of this function can be requested from the first author by e-mail.}

In Sections 5.1 and 5.2, we show and discuss the three best trees meeting the two criteria discussed in the previous sections. In Section 5.3, we present the results from the ensembles.

\subsection{The three best individual trees}\label{subsec:5.1}

\begin{figure}[H]
\centering
\includegraphics[angle=90,scale=0.5]{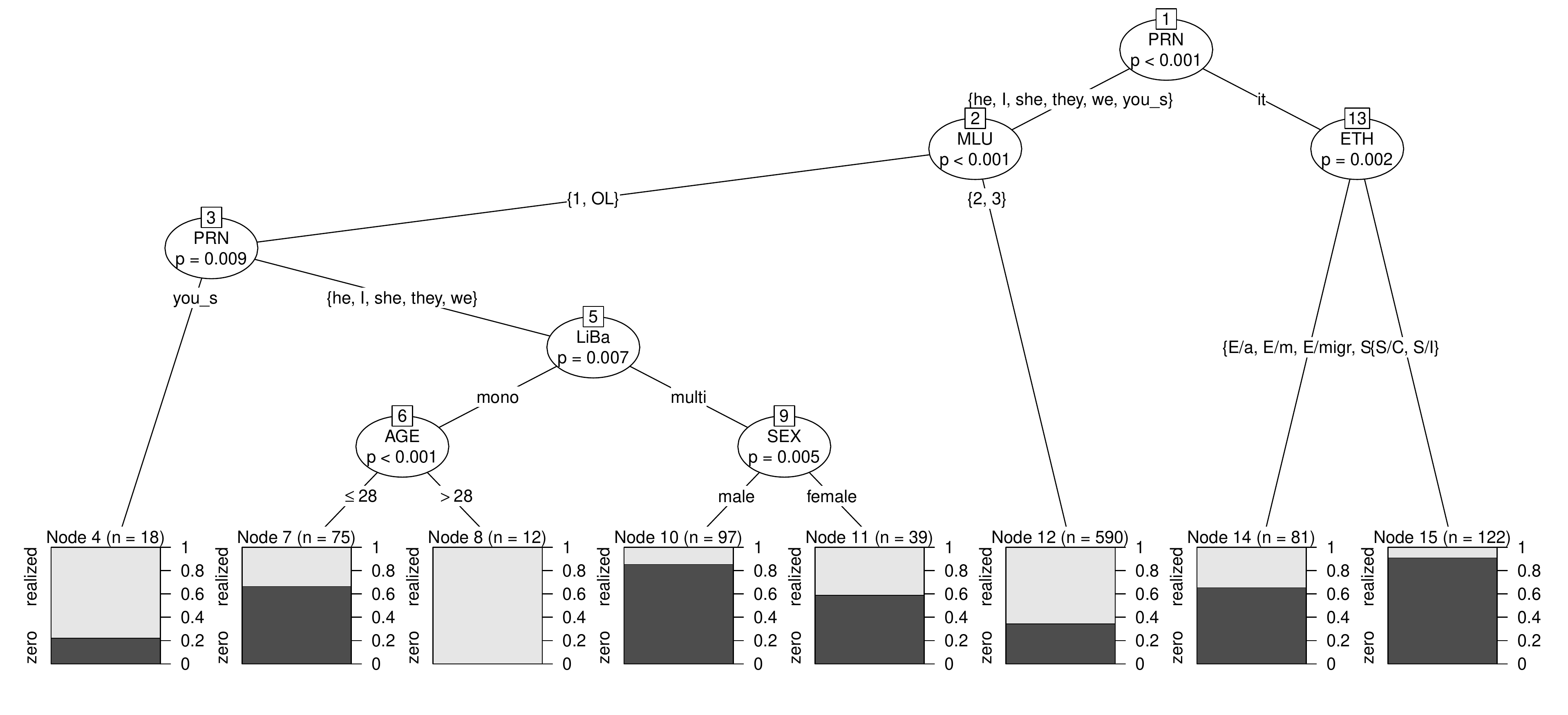}\\
\caption{Best interpretable tree from undersampling.}
\label{fig:best}
\end{figure}

\begin{figure}[H]
\centering
\includegraphics[angle=90,scale=0.48]{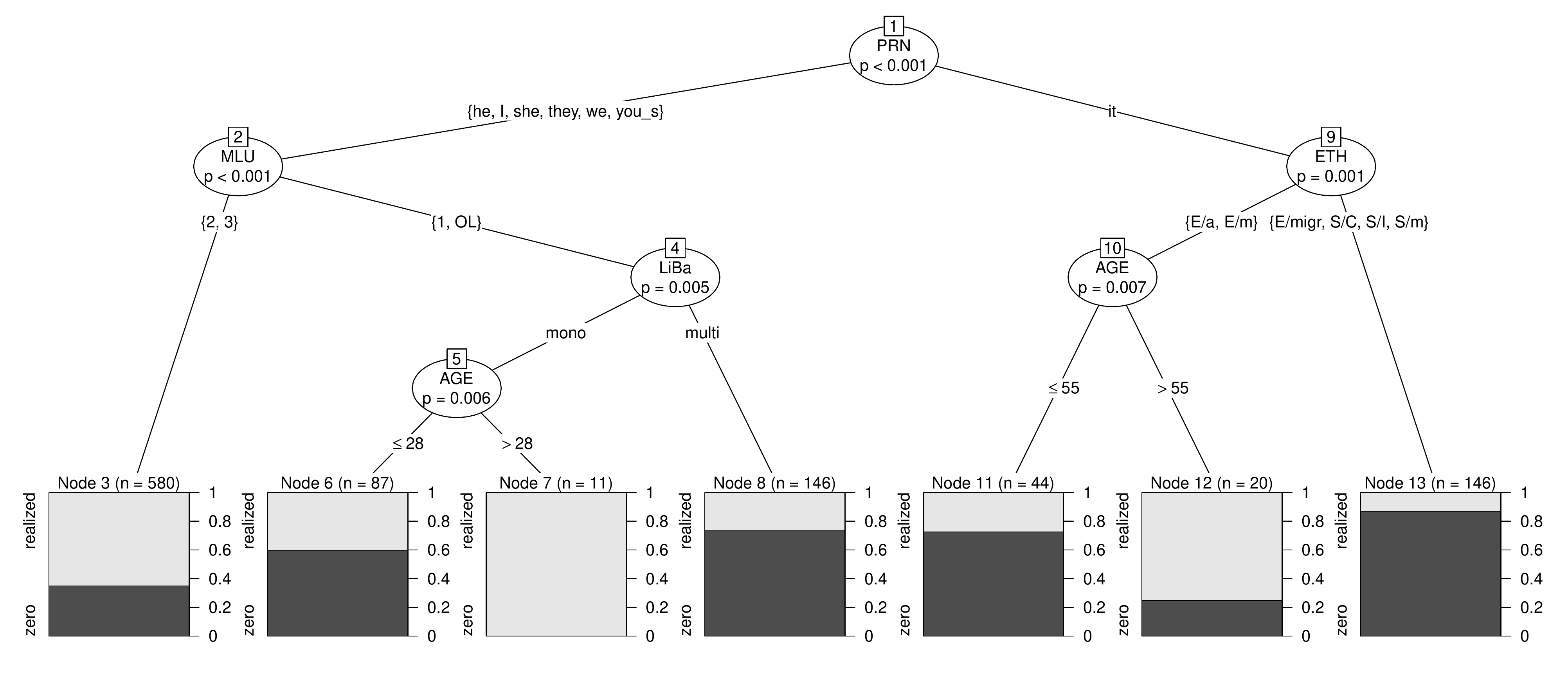}\\
\caption{Second-best interpretable tree from undersampling.}
\label{fig:best2}
\end{figure}

\begin{figure}[H]
\centering
\includegraphics[angle=90,scale=0.48]{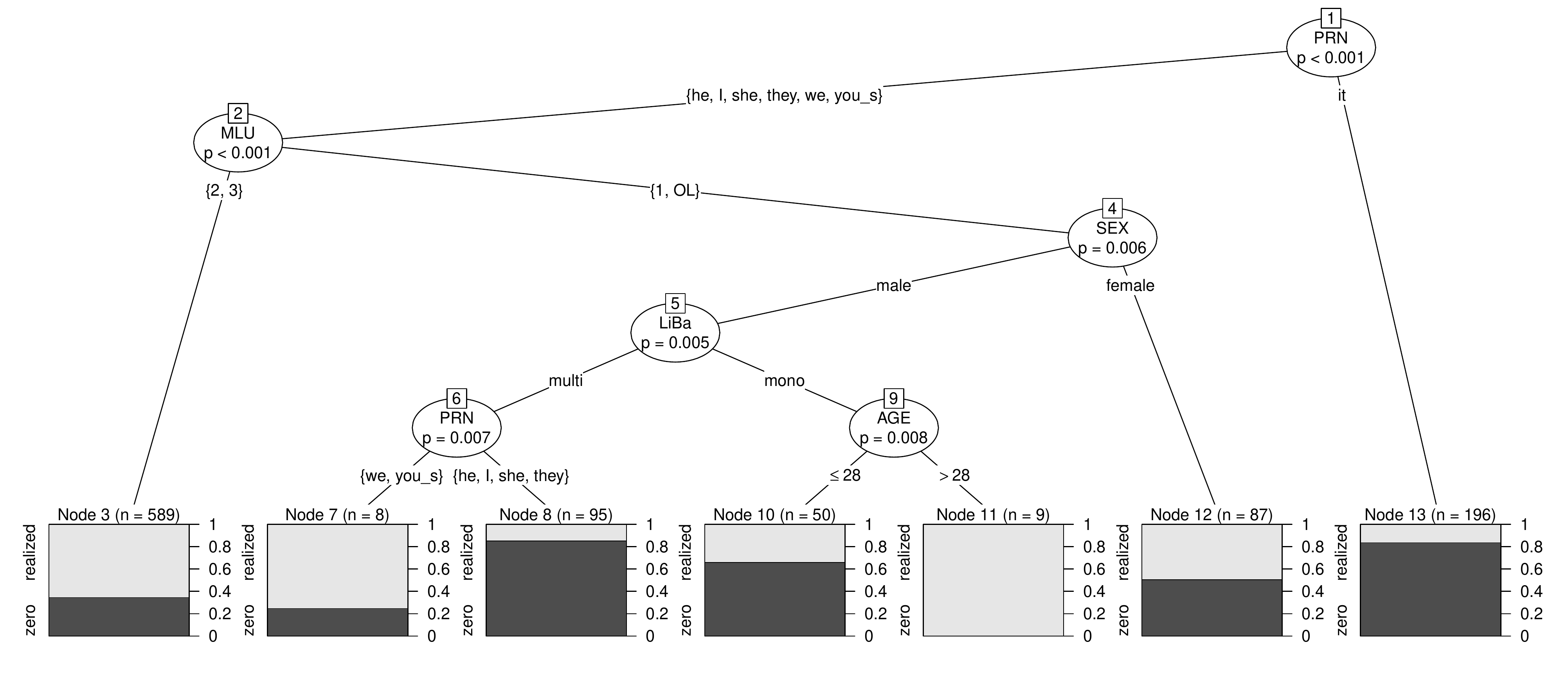}\\
\caption{Third-best interpretable tree from undersampling.}
\label{fig:best3}
\end{figure}

The three best trees presented in Figures~\ref{fig:best}--\ref{fig:best3} have balanced accuracies of 0.702, 0.701, and 0.700, respectively. We argue that, for a linguistic study, this is at least an acceptable if not high accuracy (cf. Winter 2020, 77, for a similar line of argumentation). Furthermore, all three trees fulfill the interpretability criterion and reveal a similar structure.

In all three trees, the most prominent split separates the pronoun \textit{it} from the rest (\textit{he, I, she, they, we, you} singular\footnote{Plural \textit{you} does not exist in the data.}). If further significant splits are modeled for \textit{it} (Figures~\ref{fig:best} and \ref{fig:best2}), the variables ETH and AGE are involved. The ETH split mainly separates the Singaporean children from the children growing up in England. The second-best tree (Figure~\ref{fig:best2}) further shows an AGE-related split (node 10) for the ancestral English and mixed-English children. Those children approximately four and a half years and younger use considerably more zero pronouns than the older ones in these groups (Figure~\ref{fig:best2}; nodes 11 and 12).

For the realization of the remaining pronouns, MLU, LiBa, and AGE play a role in all three trees. MLU is the most prominent variable for the rest of the pronouns and splits the children into the outliers (OL)\footnote{Four of the children in the Singapore cohort were identified as outliers on the basis of the MLU-categorization since the syntactic complexity of their utterances does not match the expected age-related performance.} and group 1, i.e. those children younger than 45 months, and the two groups of older children (MLU groups 2 and 3). In all three trees, no further splits occur for groups 2 and 3. The group 1 and OL children are split into further subgroups by LiBa. ETH determines significant splits of the data in the best and second-best tree (Figures~\ref{fig:best} and \ref{fig:best2}) and SEX has a significant impact in the best and third-best tree (Figure~\ref{fig:best} and \ref{fig:best3}).

Looking into the extreme frequencies in the manifestations of our dependent variable, the trees reveal the following: In the best tree (Figure~\ref{fig:best}), the highest share of zero pronouns can be found for multilingual male children (node 10). In the second-best tree (Figure~\ref{fig:best2}), all multilingual children show a particularly high share of zero pronouns (node 8). The third-best tree (Figure~\ref{fig:best3}) again shows that male gender and multilingual linguistic background are predictors for high rates of zero pronouns, but only for the pronouns \textit{he, I, she}, and \textit{they} (node 8).

Extreme frequencies of realized subject pronouns can be found for the following sub-rules. In the best tree (Figure~\ref{fig:best}), monolingual children older than 28 months exclusively use the realized variant (node 8). Furthermore, the realized variant clearly dominates for pronoun \textit{you} singular (node 4).

\noindent In the second-best tree (Figure~\ref{fig:best2}), again, monolingual children older than 28 months exclusively use the realized variant (node 7). MLU groups 2 and 3 also clearly favor the realized variant (node 3). In the third-best tree (Figure~\ref{fig:best3}), those male children that are multilingual use exceptionally high proportions of realized \textit{we} and \textit{you} singular (node 7) and again, MLU groups 2 and 3 make exceptionally high use of the realized variant (node 3).

\subsection{Discussion of trees}\label{subsec:5.2}

To summarize, the three best trees from our {\bf PrInDT} analysis show that all considered variables (ETH, AGE, SEX, LiBa, MLU, PRN) have an effect on the data to different degrees and in different combinations of sub-rules. The intralinguistic variable PRN has the strongest effect on the results in all three trees. Zero pronouns most prominently occur for the pronoun \textit{it}, this means either for the semantically empty dummy \textit{it} ("It is raining") or the referential \textit{it} ("My cat, it is black").

MLU, LiBa, and AGE play a role in all three trees in very similar ways: MLU always determines a split into group 1 and the outliers on the one hand and the groups 2 and 3 on the other, with the former being more prone to using zero forms than the latter.
AGE plays a different role for the pronoun \textit{it} and the group of all other pronouns. For \textit{it}, AGE splits the English ancestral and mixed children at 55 months. For the rest of the pronouns and the OL and young children (group 1), AGE splits the children at 28 months. Therefore, hypothesis 1 is confirmed: AGE and MLU are strong predictors for the realization of subject pronouns.

Hypothesis 2 is also confirmed: ETH clearly guides the realization of subject pronouns in that Singaporean children make significantly stronger use of zero subjects throughout the age groups (at least in two of the three best trees).

Further findings that go beyond our initial hypotheses have revealed that SEX has a significant impact on the realization of subject pronouns in the best and the third-best trees, with the male children showing a stronger inclination towards using the zero variant than the female children. This is not surprising either, as it is a common finding in sociolinguistic research that women often use more standard speech forms than men.

As we have seen, the trees not only have satisfactory if not high balanced accuracies, they are also fully interpretable and consistent in their linguistic findings. This nicely illustrates the methodological power of our {\bf PrInDT} approach. Note that a clear-cut threshold for linguistic studies does not exist and depends on the individual data set (cf. Section~\ref{subsec:5.3}).

\subsection{Ensembles of decision trees}\label{subsec:5.3}

\begin{figure}[H]
\centering
\vspace{-0.9cm}
\includegraphics[width=0.75\textwidth]{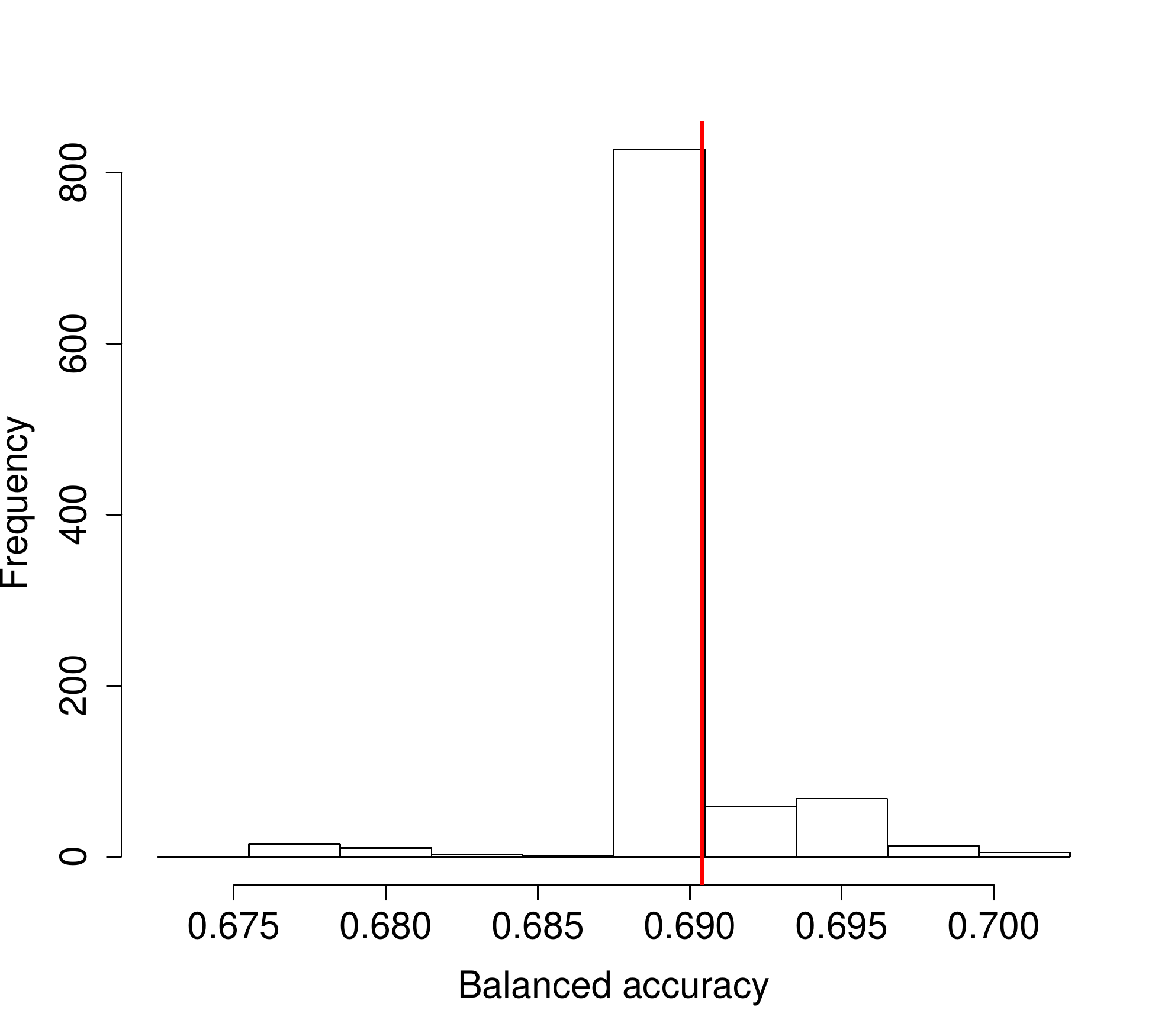}\\
\vspace{-0.3cm}
\caption{Histogram of balanced accuracies from undersampling; the median is represented by the red vertical line.}
\label{fig:ba}
\end{figure}

Let us now consider four possible ensembles of interpretable trees with acceptable predictive power.\footnote{Note that random forests, as often used in linguistics, are another type of ensembles.} Figure~\ref{fig:ba} illustrates the distribution of the balanced accuracies of all 1001 trees from undersampling. Note that the range of balanced accuracies is quite small (from 0.6755 to 0.7021). In particular, this means that none of the trees has a totally unacceptable predictive power.

\vspace{-0.2cm}
\noindent In this paper, we consider
\begin{itemize}
\itemsep0pt
\item[a)] the ensemble comprising only the three interpretable trees with the highest balanced accuracies (discussed in Section~\ref{subsec:5.1}),
\item[b)] the ensemble consisting of {\bf all} interpretable trees from undersampling,
\item[c)] the ensemble including all interpretable trees with a predictive power greater than the median of the balanced accuracies of all trees from undersampling (c = 0.6904; cf. Figure~\ref{fig:ba}).
\end{itemize}

\begin{table}[H]
\centering
\caption{No. of trees and balanced accuracies for the different ensembles}
\label{tab:acc}
\vspace{0.3cm}
    \begin{tabular}{l|rr}
    ensemble & trees & accuracy \\
    \hline
    a) & 3   & 0.699\\
    b) & 941 & 0.690\\
    c) & 137 & 0.696
  \end{tabular}
\end{table}

\noindent Table~\ref{tab:acc} summarizes the ensembles and their respective accuracies. Note that only 60 of the 1001 trees turned out to be uninterpretable (cf. b)). Overall, ensemble a), which comprises the three interpretable trees with the best balanced accuracies, comes with the {\bf best balanced accuracy} (0.699) of the ensembles.
The second-best ensemble with an accuracy of 0.696 is ensemble c), which consists of 137 trees, which have greater balanced accuracies than the median of all balanced accuracies.
Ensemble b), which consists of all interpretable trees from undersampling, includes 
941 trees and has the lowest accuracy of the ensembles (0.690). Still, its accuracy is close to the median. 

\subsection{Discussion of ensembles}

Finally, we would like to argue that ensembles can add important information to the findings of individual trees. Small samples in resampling represent only small parts of the data set. Therefore, trees based on these samples also only depict parts of the reality. Ensembles ensure that as many of these parts are considered as determined by the number of repetitions (in our example 3, 941, and 137). It is further interesting to investigate how strongly the accuracies vary when the underlying samples are changing as this gives some indication of the robustness of the findings.

From a linguistic perspective, working with ensembles makes sense for three reasons. Similar to the first statistical reason outlined above, working with trees trained on resampled data sets runs the risk of neglecting important parts of the data. Ensembles are more representative of the data set and thus of the overall population. Furthermore, randomly selected ensembles mirror the randomness of the data collected for a linguistic study. In both cases, only parts of the overall population can be captured with basically everybody or every token having an equal chance of being selected. Moreover, ensembles account for the nature of sociolinguistic variation, since it is neither fully random nor deterministic. As the last 60 years of sociolinguistic research have clearly shown, linguistic variation is guided by intra- and extralinguistic factors but at the same time shows inter- and even intraspeaker variation (e.g. Meyerhoff 2019, 12). Our study once more confirms these findings.
Ensembles, therefore, aptly represent the strong systematicity of linguistic data on the one hand while at the same time leaving room for randomness and variation.
The one shortcoming of ensembles for linguistic studies, however, is that the interpretability of the model is blurred since it is difficult to synthesize the interpretation of a high number of trees.

\section{Conclusion}

In this paper, we have seen that statistical models and their evaluation should be integral parts of linguistic analyses, but only if the following crucial points are taken into consideration:
\begin{itemize}
\itemsep0pt
\item[1.]	 Perfectly interpretable trees with allegedly clear and strong linguistic findings can be misleading due to extremely low predictive power (see Figure~\ref{fig:simple}).
\item[2.]	Trees with high accuracies can be linguistically uninterpretable (see Figure~\ref{fig:uninter}).
\end{itemize}
To study and illustrate this finding, we have developed and presented the statistical method {\bf PrInDT} ({\bf Pr}ediction and {\bf In}terpretation with {\bf D}ecision {\bf T}rees),
which prescribes interpretability and high predictive power as properties for trees generated by resampling. Either the individual interpretable trees with highest predictive power or a whole ensemble of interpretable trees with `high enough' predictive power are then used for the prediction of relationships between linguistic variables. We successfully applied this method to find prediction rules for the use of subject pronouns (realized vs. zero) depending on extra- and intralinguistic variables.

We would like to conclude with some general remarks on statistical modeling, not only but in particular in linguistics:
\begin{description}
\itemsep0pt
\item[\bf Predictive power:] {\bf Bad models can never be interpreted.} Therefore, evaluation is a crucial step of model building. Significant splits alone do not qualify a model for usage. Interpretation of models with a bad fit or predictive power intrinsically carries the risk of drawing the wrong conclusions.
\item[\bf Interpretability of trees and ensembles:] {\bf Use interpretable models only.} Restricting models by means of prior knowledge, i.e.\ findings from earlier studies, or on the basis of theoretical assumptions enhances the interpretability of trees and the homogeneity of ensembles. This way, all trees are in accordance with prior knowledge or theory. If, despite of this, an ensemble turns out to have low predictive power, then our data do not match the underlying theoretical assumptions and the model cannot be interpreted (see Predictive power above).
\end{description}

\noindent Last but not least, we would like to stress that our method {\bf PrInDT} is not restricted to undersampling but can be applied to any ensemble, e.g. to random forests and other more general resampling methods. Therefore, as a next step we would also like to withhold (small) parts of the smaller class from the training of the trees. This way, the smaller class is also predicted for some tokens. Moreover, in order to facilitate the interpretation of ensembles, we will implement a ranking of the importance of predictors in ensembles, similar to the ranking of variables in random forests.

\section{References}
\begin{hangparas}{.25in}{1}
Buschfeld, S. (2020), `Children's English in Singapore: Acquisition, Properties, and Use', Routledge.

Chan, M. (2020), `English, mother tongue and the Singapore identity', The Straits Times, url: https://www.straitstimes.com/opinion/english-mother-tongue-and-the-spore-identity.

Gries, S.Th. (2020), `On classification trees and random forests in corpus linguistics: Some words of caution and suggestions for improvement', Corpus Linguistics and Ling. Theory 16(3): 617--647.

Meyerhoff, M. (2019), `Introducing Sociolinguistics', 3rd edn; Routledge.

R Core Team (2019), `R: A language and environment for statistical
  computing'. R Foundation for Statistical Computing, https://www.R-project.org/.

Roeper, T., Rohrbacher, B. (2000). `Null subjects in early child language and the economy of projection'. In S. Powers, C. Hamann (Eds.), Acquisition of Scrambling and Cliticization, 345--397. Berlin: Springer.

Tagliamonte, S.A., Baayen, R.H. (2012), `Models, forests, and trees of York English: Was/were variation as a case study for statistical practice', Language Variation and Change 24, 135--178.

Weiss, G.M. (2004), `Mining with rarity: A unifying framework', ACM SIGKDD Explorations 6: 7--19.

Winter, B. (2020), `Statistics for Linguists. An Introduction using R', Routledge.
\end{hangparas}

\end{document}